\documentclass[runningheads]{llncs}
\usepackage[T1]{fontenc}
\usepackage{lineno}

\usepackage{graphicx,verbatim}
\usepackage{tabularx}
\newcolumntype{Y}{>{\raggedright\arraybackslash}X}
\usepackage{hyperref}
\usepackage{amsmath}
\usepackage{makecell}
\usepackage{amssymb}
\usepackage{amsmath}

\usepackage{multirow}
\usepackage{pifont}
\usepackage{booktabs}
\usepackage{tabularx}
\usepackage{array}
\usepackage[misc]{ifsym}

\usepackage{arydshln}
\usepackage{tablefootnote}
\usepackage{multicol}

\newcolumntype{C}{>{\hspace{0.5em}}c<{\hspace{0.5em}}}

\usepackage{xcolor}
\definecolor{blue1}{RGB}{220,234,247}
\definecolor{green1}{RGB}{217,242,208}
\definecolor{yellow1}{RGB}{255,255,204}
\usepackage{layout}
\usepackage{adjustbox}

\usepackage{color}

\begin{document}

\title{Training-free Test-time Improvement for Explainable Medical Image Classification}
\titlerunning{Training-free Test-time Improvement for Medical CBMs}

\author{Hangzhou He\inst{1,2,3,5}%
\and%
Jiachen Tang\inst{4,5}%
\and%
Lei Zhu\inst{1,2,3}%
\and%
Kaiwen Li\inst{1,2,3}%
\and%
\\Yanye Lu\inst{1,2,3,6}%
}%
\authorrunning{H. He et al.}
\institute{Department of Biomedical Engineering, Peking University, Beijing, China \and
Institute of Medical Technology, Peking University, Beijing, China \and
National Biomedical Imaging Center, Peking University, Beijing, China \and
Beijing Institute of Technology, Beijing, China \and
Equal Contribution \and
Corresponding Author \\
\email{zhuang@stu.pku.edu.cn, yanye.lu@pku.edu.cn}}

\maketitle
\vspace*{-0.3cm}
\begingroup
\renewcommand\thefootnote{}\footnotetext{
This is the initial version of our work accepted by MICCAI 2025. We'll include a link to the version on SpringerLink after this becomes available.
}
\addtocounter{footnote}{-1}
\endgroup
\begin{abstract}
Deep learning-based medical image classification techniques are rapidly advancing in medical image analysis, making it crucial to develop accurate and trustworthy models that can be efficiently deployed across diverse clinical scenarios. Concept Bottleneck Models (CBMs), which first predict a set of explainable concepts from images and then perform classification based on these concepts, are increasingly being adopted for explainable medical image classification. However, the inherent explainability of CBMs introduces new challenges when deploying trained models to new environments. Variations in imaging protocols and staining methods may induce concept-level shifts, such as alterations in color distribution and scale. Furthermore, since CBM training requires explicit concept annotations, fine-tuning models solely with image-level labels could compromise concept prediction accuracy and faithfulness - a critical limitation given the high cost of acquiring expert-annotated concept labels in medical domains. To address these challenges, we propose a training-free confusion concept identification strategy. By leveraging minimal new data (e.g., 4 images per class) with only image-level labels, our approach enhances out-of-domain performance without sacrificing source domain accuracy through two key operations: masking misactivated confounding concepts and amplifying under-activated discriminative concepts. The efficacy of our method is validated on both skin and white blood cell images. Our code is available at: \url{https://github.com/riverback/TF-TTI-XMed}.


\end{abstract}
\section{Introduction}
\begin{figure}[!t]
    \centering
    \includegraphics[width=\linewidth]{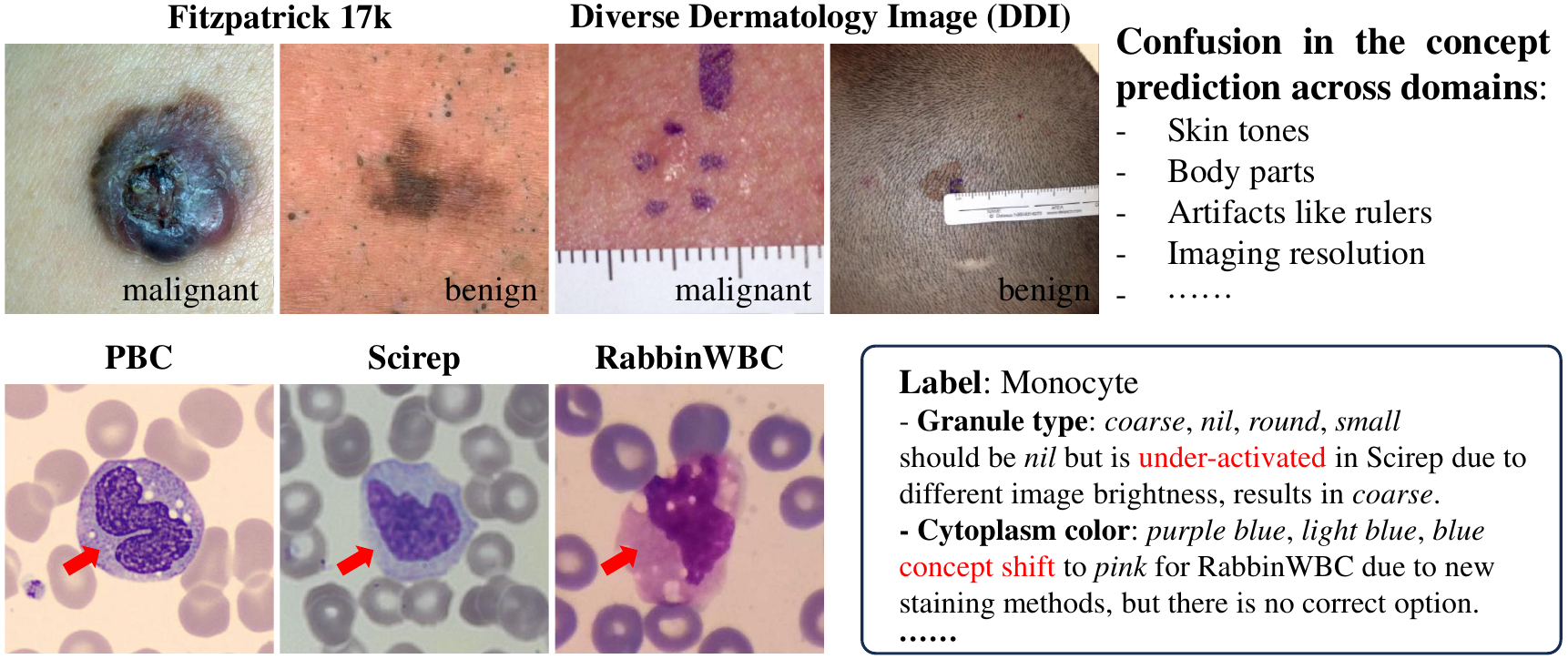}
    \caption{Challenges when deploying CBMs to new environments: the concept features across different domains may vary intensively and this may result in under-activated discriminative concepts like the \textit{nil granule type} for monocytes, or shifted concepts like the \textit{pink cytoplasm color} in RabbinWBC (which are mostly blue tones in the PBC dataset). However, annotating fine-grained concept labels for the new data is costly due to the expertise, and directly fine-tuning with image-level labels may lead to over-fitting and degrade source domain performance.}
    \label{datset_introduction_fig1}
\end{figure}
The development and integration of deep learning technologies have greatly transformed the field of medical image analysis and achieved remarkable progress in multiple medical fields~\cite{acosta2022multimodal,moor2023foundation,Varghese2024aiinsurgery,kang2025aiindrug}. However, the particularity of medical applications poses additional challenges to the explainability of deep learning models, such as checking whether the models are reliable~\cite{fajtl2024trustworthy}, fair~\cite{chen2023algorithmic}, and robust~\cite{balendran2025scoping}. However, for a long time, there has been a trade-off between model explainability and accuracy in the field of deep learning, and the accuracy of explainable models is often significantly lower than that of their black-box counterparts~\cite{gunning2019darpa}, which limits their application in real-world medical scenarios where accuracy is prioritized~\cite{accuracy_not_fairness_NEJM}. As an emerging self-explainable model architecture, the concept bottleneck model (CBM~\cite{DBLP:conf/icml/KohNTMPKL20}) uses a set of interpretable concepts as the intermediate representation of images and then relies on an interpretable linear classification layer to achieve interpretable image classification. When an accurate and sufficient concept set is provided, CBMs can often maintain good explainability and achieve performance comparable to that of black-box models and thus have been widely applied to medical image classification tasks~\cite{DBLP:conf/miccai/ChowdhuryPLTXHVL24,DBLP:conf/miccai/KimKCK23,DBLP:conf/miccai/PangKTW24,DBLP:conf/nips/0006GWWYCGY24}.

However, deploying CBMs to a new environment while maintaining explainability is challenging, e.g., a new medical institution or imaging device. Firstly, the concept features across domains may vary intensively (as shown in Fig.~\ref{datset_introduction_fig1}), which poses challenges for both fine-tuning in the target domain and accurate concept prediction, and the clinical knowledge required for annotating fine-grained medical concepts further exacerbates this drawback. Although some label-free methods have been proposed to construct CBMs by leveraging the capabilities of large language models or vision-language models~\cite{he2025v2c,DBLP:conf/iclr/OikarinenDNW23,DBLP:conf/cvpr/YangPZJCY23}, it is difficult for us to assess whether the generated medical concepts are accurate and faithful~\cite{huang2024concept}. Secondly, even if there is concept-level labeled data in the target domain, unless sufficient, direct fine-tuning of the model may lead to severe over-fitting and also degrade the model's performance on the source domain~\cite{mccloskey1989catastrophic}. Moreover, fine-tuning with only image labels may also undermine the faithfulness of CBMs, as the model may forget previously learned knowledge and fail to leverage the desired concepts.~\cite{DBLP:conf/nips/MaiC00CPB0S0C24}.

To address the above issues, we propose a training-free \textbf{confusion concept identification strategy} to improve the test-time performance of CBMs. By identifying and adjusting confusing concepts during the test phase, we can \textbf{mask misactivated confounding concepts} or \textbf{amplify under-activated discriminative concepts}. More importantly, our method does not harm the faithfulness of the concept prediction procedure and in-domain performance. Experimental results on skin and white blood cell images demonstrate that, with minimal data with only image labels (e.g., only 4 images per class), our method achieves performance comparable to fine-tuning, without sacrificing the source domain's accuracy while keeping the explainability of concept bottleneck models.

\section{Related Work}
\subsection{Concept Bottleneck Models for Medical Image Classification}
The concept bottleneck model (CBM~\cite{DBLP:conf/icml/KohNTMPKL20}) maps visual images to a space composed of predefined concepts, extracts intermediate concepts as the representation of the images, and uses an interpretable linear classifier for classification. Due to its inherent explainability, CBMs have been extensively utilized in medical image classification tasks, including white blood cell classification~\cite{DBLP:conf/miccai/PangKTW24}, skin disease classification~\cite{DBLP:conf/miccai/KimKCK23}, fundus disease classification~\cite{DBLP:conf/miccai/ChowdhuryPLTXHVL24}, and the classification of thorax diseases~\cite{DBLP:conf/nips/0006GWWYCGY24}. Among these works, Pang \textit{et al.} proposed to align CBMs with clinical knowledge during training to improve out-of-domain (OOD) performance, but needed additional clinical knowledge labels to indicate concept preference~\cite{DBLP:conf/miccai/PangKTW24}. Yang \textit{et al.} proposed Knowledge-enhanced Bottlenecks that used language models and PubMed as the knowledge resources for making CBMs less sensitive to domain shift~\cite{DBLP:conf/nips/0006GWWYCGY24}. Nevertheless, existing methods for improving the OOD performance of CBMs all require additional training or the integration of a vision-language model like CLIP to help concept prediction at test time~\cite{choi2025adaptive}, which can be challenging to quickly deploy in new environments. 

Beyond these methods, CBM itself is also renowned for its test-time concept intervention capability, which permits users to modify activation values of specific concepts at the concept bottleneck~\cite{DBLP:conf/icml/ShinJAL23}. However, when applied to OOD medical images, such interventions often rely on specialized medical knowledge and fail to consistently improve model performance across diverse datasets, thereby limiting their generalizability and practical utility. In response to this challenge, the Concept Bottleneck Memory Model (CB2M)~\cite{DBLP:conf/icml/SteinmannSFK24} introduces a two-fold memory module designed to capture model errors during training and validation phases, utilizing these memories to bolster test performance. Nonetheless, CB2M fails to adequately explore the issue of concept shift in OOD scenarios, which remains a critical concern. Another prominent approach for adapting pre-trained models to distribution shifts is test-time adaptation (TTA)~\cite{DBLP:journals/ijcv/LiangHT25}. TTA methods typically exploit unlabeled test-domain data or data streams to facilitate domain adaptation. However, the two-step inference process of CBM necessitates the preservation of the model's concept prediction fidelity during the TTA phase. Regrettably, research on enhancing CBM's test-time performance while maintaining fidelity remains unexplored.

\section{Method}

\subsection{Preliminary: Concept Bottleneck Model\label{preliminary}}
\begin{figure}[!t]
\includegraphics[width=\textwidth]{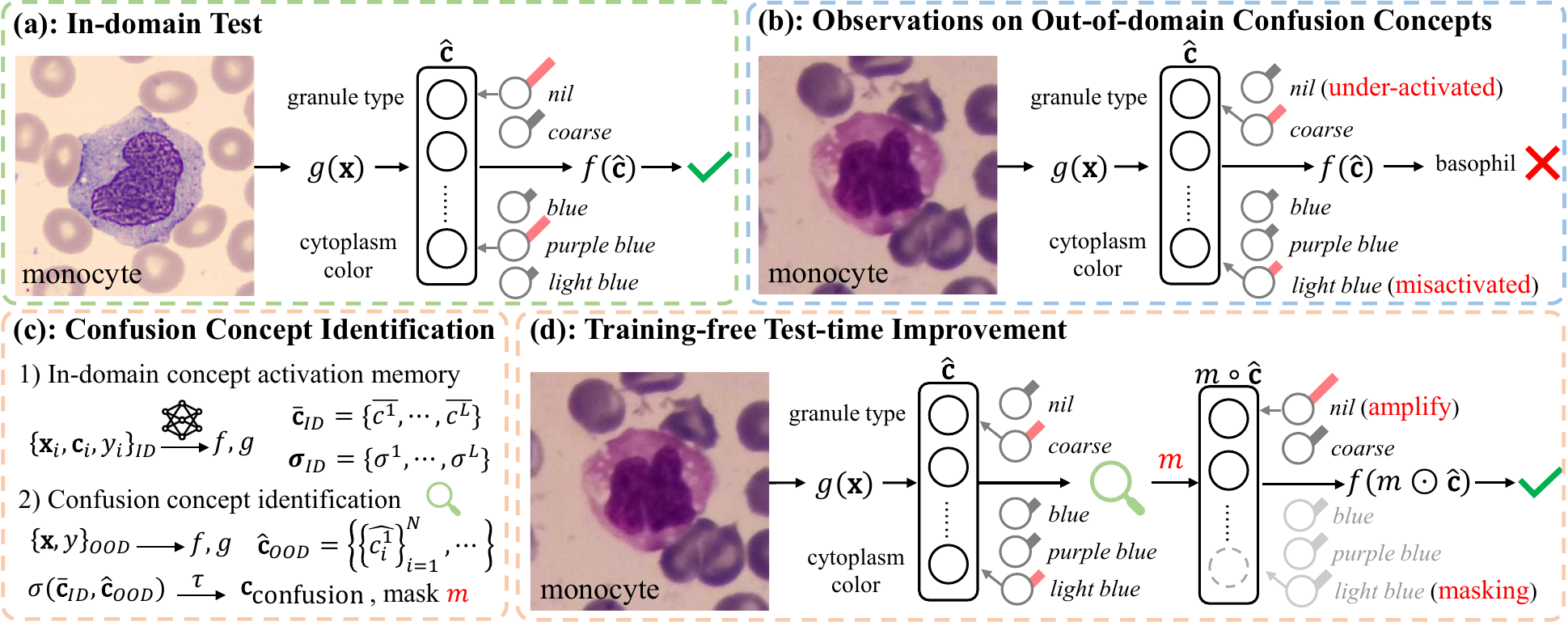}
\caption{Method overview: (\textbf{a}) a trained CBM can perform explainable classification with accurate concepts, (\textbf{b}) but fails on the out-of-domain (OOD) data because of two types of confusion concepts, under-activated discriminative concepts and misactivated confusing concepts. So we (\textbf{c}) propose a confusion concept identification strategy and (\textbf{d}) use the concept mask $m$ to amplify or mask these concepts.} \label{method_overview}
\end{figure}

Formally, consider an image dataset ${\cal D}=\{(\textbf{x}_i, \textbf{c}_i, y_i)\}$ where $\textbf{x}$ denotes an image, $\textbf{c}=\{c^1, c^2,\cdots, c^L\}$ is $L$ predefined concepts, and $y\in \{0,1\}^{K\times 1}$ is the image label from $K$ target classes. As shown in Fig.~\ref{method_overview} (a), a CBM first predicts concepts using a concept predictor $g$ and then predicts targets based on the concepts using a classifier $f$: 
\begin{equation}
    \hat{\textbf{c}} = g(\textbf{x}),\ \hat{y} = f(\hat{\textbf{c}})
\end{equation}
In this work, we jointly train $g$ and $f$ to make the framework compatible with additional optimization strategies such as incorporating clinical knowledge~\cite{DBLP:conf/icml/ShinJAL23}.
\begin{equation}
    \hat{f}, \hat{g} = \mathop{\arg\min}\limits_{f,g} \sum_i^{|{\cal D}|} \left[ \varphi {\cal L}_c (g(\textbf{x}_i); \textbf{c}_i) + {\cal L}_y (f(g(\textbf{x}_i));y_i) \right]
    \label{loss_function}
\end{equation}
where ${\cal L}_c$ and ${\cal L}_y$ are loss functions for concept prediction and class prediction respectively, for some $\varphi > 0$.

\subsection{Confusion Concept Identification Strategy\label{strategy}}
We first give some observations into the concept prediction procedure on OOD data. As shown in Fig.~\ref{method_overview} (b), there are typically two types of confusion concepts: \textbf{1) under-activated discriminative concepts} that are crucial for predicting the target class but are less activated than other concepts because of feature distribution shifts caused by different image brightness, resolution and so on; and \textbf{2) misactivated confounding concepts concepts} that are unable to be faithfully activated in the OOD dataset due to huge changes in feature representation, like the blue tones cytoplasm color is changed to pink in the RabbinWBC dataset. Based on these observations, we propose our \textbf{confusion concept identification strategy} which leverages the concept activation memory during the in-domain (ID) model development and then uses minimal image-level labeled data from the target domain to identify the two types of confusion concepts. As presented in Fig.~\ref{method_overview} (c), we save the mean ({$\overline{\textbf{c}}_{ID}$}) and the standard deviation ({$\boldsymbol \sigma_{ID}$}) for concept activation value on the validation set ${\cal D}_{val}$ as the concept activation memory for each target class $k$:
\begin{equation}
    \{\overline{\textbf{c}}_{ID},{ \boldsymbol \sigma }_{ID}\}_k = \{(\overline{c^1},\sigma^1), \cdots, (\overline{c^L}, \sigma^L) \}_k, \ k\in \{1, 2, \cdots, K\}
\end{equation}
where the subscript $k$ denotes the $k$-th element of a set. For example, the concept activation memory of class $k$ for $c^1$ is calculated as:
\begin{equation}
    \overline{c^1} = \frac{1}{|{\cal D}_{val}(k)|} \sum_{i,y_i=k}^{|{\cal D}_{val}(k)|} g(\textbf{x}_i)[0],\ \sigma^1 = \sqrt{\sum_{i,y_i=k}^{|{\cal D}_{val}(k)|} \frac{1}{|{\cal D}_{val}(k)|} (g(\textbf{x}_i)[0]-\overline{c^1})^2} 
\end{equation}
Given a set of image-level labeled OOD images ${\cal D}_{OOD}=\{\textbf{x}_i,y_i\}^{N\times K}_{i=1}$ where $N$ is the number of images for each class and is usually small, we iterate these data once to get the concept activation information in the target domain. The OOD concept activation information for class $k$ is:
\begin{equation}
   \widehat{\textbf{c}}_{OOD,k} = \left\{\{\widehat{c^1_i}\}^N_{i=1}, \cdots, \{\widehat{c^L_i}\}^N_{i=1}\right \}_k
\end{equation}
then we can identify confusing concepts for class $k$ by calculating the stability of concept activation (e.g., standard deviation) between ID and OOD samples:
\begin{equation}
    \textbf{c}_{{\rm confusion},k} =  \bigcup_{i=1,\cdots,L} \left \{ {c^i} \ \bigg| \ \sigma [{\rm concat}(\ \overline{\textbf{c}}_{ID,k}[i],  \widehat{\textbf{c}}_{OOD,k}[i]\ )] \geq \tau \cdot {\boldsymbol \sigma}_{ID,k}[i]\right \} \label{eq6}
\end{equation}
where $\sigma$ denotes the computation of the standard deviation, the factor $\tau>1$ ensures the OOD concept activation variation is larger than that of ID. We use $\tau=1.4$ for skin and $\tau=1.1$ for WBC. A concept is deemed misactivated if it confuses more than half of the classes. The remaining confusing concepts are treated as under-activated discriminative concepts if their in-domain activation probability for the corresponding class exceeds 0.5. 

\subsection{Confusion Concepts Activation Manipulation\label{operations}}
Upon acquiring the two categories of confusion concepts, we can adjust the activation levels of these concepts to enhance test-time performance. Towards this end, we put forward two straightforward yet efficacious operations: \textbf{(1) Masking Misactivated Confounding Concepts}, and \textbf{(2) Amplifying Under-Activated Discriminative Concepts}. As depicted in Fig.~\ref{method_overview} (d), we initialize a concept mask $m$ with a value of 1. We mask the misactivated confounding concepts by setting the corresponding values to zero. Meanwhile, to amplify the under-activated discriminative concepts, we multiply them by a factor greater than 1; for the sake of simplicity, we adopt 2 in this paper. During inference, we simply multiply the concept mask $m$ element-wise with the predicted concepts $\hat{\textbf{c}}$, and use the result as input to the final classifier $f$:
\begin{equation}
    \hat{y}_{\rm new} = f(m \odot \hat{\textbf{c}}) = f(m \odot g(\textbf{x}))
\end{equation}

\section{Experiment}

\begin{table}[t]
\centering
\caption{Classification accuracy and F1 score on skin images. ${\cal F}$ means fine-tuning, ${\cal K}$ means integrating clinical knowledge, and N is the number of image-level labeled images for each class in the out-of-domain dataset. We use the same images for ${\cal F}$ and our method.
}
\label{tab:classification_acc_skin}

\begin{tabularx}{\textwidth}{YYYYY}
\toprule

& \multicolumn{2}{c}{\textbf{Fitzpatrick 17k (In-domain)}} & \multicolumn{2}{c}{\textbf{DDI (Out-of-domain)}}  \\
\cmidrule(lr){2-3} \cmidrule(lr){4-5}
\multirow{-3}{*}{\textbf{\begin{tabular}[c]{@{}c@{}}CBM\\Backbone\end{tabular}}} & Accuracy & F1 Score & Accuracy & F1 Score \\
\midrule
VGG16
& $75.28$   & $75.28$   & $65.72$   & $58.72$ \\
+${\cal F}$ (N=8)
& $61.40_{\textsubscript{$\downarrow 13.88$}}$   & $57.09_{\textsubscript{$\downarrow 17.38$}}$   & $69.65_{\textcolor{red}{\textsubscript{$\uparrow 3.93$}}}$   & $56.58_{\textsubscript{$\downarrow 2.14$}}$ \\
+Ours (N=4)
& $74.53_{\textsubscript{$\downarrow 0.75$}}$   & $74.36_{\textsubscript{$\downarrow 0.92$}}$   & $\textbf{70.28}_{\textcolor{red}{\textsubscript{$\uparrow 4.56$}}}$   & $\textbf{60.46}_{\textcolor{red}{\textsubscript{$\uparrow 1.74$}}}$ \\
+Ours (N=8)
& $74.42_{\textsubscript{$\downarrow 0.86$}}$   & $74.25_{\textsubscript{$\downarrow 1.03$}}$   & $70.44_{\textcolor{red}{\textsubscript{$\uparrow 4.72$}}}$   & $60.75_{\textcolor{red}{\textsubscript{$\uparrow 2.03$}}}$ \\
\hdashline
VGG16+${\cal K}$
& $76.95$   & $76.93$   & $62.11$   & $56.00$ \\
+${\cal F}$ (N=8)
& $64.79_{\textsubscript{$\downarrow 12.16$}}$   & $63.43_{\textsubscript{$\downarrow 13.50$}}$   & $62.11$   & $59.44_{\textcolor{red}{\textsubscript{$\uparrow 3.44$}}}$ \\
+Ours (N=4)
& $76.95$   & $76.94_{\textcolor{red}{\textsubscript{$\uparrow 0.01$}}}$   & $66.19_{\textcolor{red}{\textsubscript{$\uparrow 4.09$}}}$   & $57.55_{\textcolor{red}{\textsubscript{$\uparrow 1.55$}}}$ \\
+Ours (N=8)
& $76.95$   & $76.95_{\textcolor{red}{\textsubscript{$\uparrow 0.02$}}}$   & $65.88_{\textcolor{red}{\textsubscript{$\uparrow 3.77$}}}$   & $57.16_{\textcolor{red}{\textsubscript{$\uparrow 1.16$}}}$ \\
\midrule
ResNet34
& $77.72$   & $77.72$   & $62.89$   & $59.09$ \\
+${\cal F}$(N=8)
& $68.61_{\textsubscript{$\downarrow 8.78$}}$   & $68.60_{\textsubscript{$\downarrow 9.12$}}$   & $64.94_{\textcolor{red}{\textsubscript{$\uparrow 2.05$}}}$   & $60.22_{\textcolor{red}{\textsubscript{$\uparrow 1.13$}}}$ \\
+Ours(N=4)
& $77.98_{\textcolor{red}{\textsubscript{$\uparrow 0.26$}}}$   & $77.97_{\textcolor{red}{\textsubscript{$\uparrow 0.25$}}}$   & $64.15_{\textcolor{red}{\textsubscript{$\uparrow 1.26$}}}$   & $60.13_{\textcolor{red}{\textsubscript{$\uparrow 1.04$}}}$ \\
+Ours(N=8)
& $77.32_{\textsubscript{$\downarrow 0.40$}}$   & $77.22_{\textsubscript{$\downarrow 0.50$}}$   & $70.13_{\textcolor{red}{\textsubscript{$\uparrow 7.23$}}}$   & $62.28_{\textcolor{red}{\textsubscript{$\uparrow 3.20$}}}$ \\
\hdashline
ResNet34+${\cal K}$
& $77.38$   & $77.29$   & $62.26$   & $58.48$ \\
+${\cal F}$ (N=8)
& $67.61_{\textsubscript{$\downarrow 9.77$}}$ & $67.61_{\textsubscript{$\downarrow 9.68$}}$ & $67.14_{\textcolor{red}{\textsubscript{$\uparrow 4.88$}}}$ & $59.17_{\textcolor{red}{\textsubscript{$\uparrow 0.69$}}}$\\
+Ours (N=4)
& $78.41_{\textcolor{red}{\textsubscript{$\uparrow 1.12$}}}$   & $78.41_{\textcolor{red}{\textsubscript{$\uparrow 1.03$}}}$   & $66.98_{\textcolor{red}{\textsubscript{$\uparrow 4.72$}}}$   & $60.48_{\textcolor{red}{\textsubscript{$\uparrow 2.00$}}}$ \\
+Ours (N=8)
& $77.84_{\textcolor{red}{\textsubscript{$\uparrow 0.46$}}}$   & $77.73_{\textcolor{red}{\textsubscript{$\uparrow 0.44$}}}$   & $68.55_{\textcolor{red}{\textsubscript{$\uparrow 6.29$}}}$   & $59.86_{\textcolor{red}{\textsubscript{$\uparrow 1.38$}}}$ \\
\bottomrule
\end{tabularx}
\end{table}

\begin{table}[tb!]
\centering
\caption{Classification accuracy and F1 score on WBC images. ${\cal F}$ means fine-tuning, ${\cal K}$ means integrating clinical knowledge, and N is the number of image-level labeled images for each class in the out-of-domain dataset. We use the same images for ${\cal F}$ and our method.
}
\label{tab:classification_acc_wbc}
\begin{adjustbox}{width=\textwidth}
\begin{tabular}{lcccccc}
\toprule
\multirow{3}{*}{\textbf{\begin{tabular}[c]{@{}c@{}}CBM \\Backbone \end{tabular}}} & \multicolumn{2}{c}{\textbf{In-domain Dataset}} & \multicolumn{4}{c}{\textbf{Out-of-domain\  Datasets}}\\
\cmidrule(lr){2-3} \cmidrule(lr){4-7} & \multicolumn{2}{c}{\textbf{PBC}${}^1$} & \multicolumn{2}{c}{\textbf{Scirep}}  & \multicolumn{2}{c}{\textbf{RaabinWBC}} \\
\cmidrule(lr){2-3} \cmidrule(lr){4-5} \cmidrule(lr){6-7} & Accuracy & F1 Score & Accuracy & F1 Score& Accuracy & F1 Score\\
\midrule
VGG16
& $99.71$ & $99.61$ & $84.10$ & $70.85$     & $75.06$     & $50.14$  \\
+${\cal F}$ (N=4)
& $97.42$/$96.16$ & \ $96.96$/$97.09$ & $88.01$ & $75.55$ & $86.89$ & $78.34$  \\
+Ours (N=4)
& $99.71$/$99.52$ & \ $99.61$/$99.41$ & $84.25$ & $71.45$ & $76.05$ & $51.64$  \\
+Ours (N=8)
& $99.32$/$99.71$ & \ $99.45$/$99.74$ & $86.43$ & $72.18$ & $78.43$ & $55.09$  \\
\hdashline
VGG16+${\cal K}$
& $99.84$ & $99.77$ & $84.00$  & $67.37$ & $77.21$ & $51.89$  \\
+${\cal F}$ (N=4)
& $91.87$/$96.45$ & \ $90.64$/$96.10$ & $84.50$ & $77.20$ & $88.20$ & $82.62$   \\
+Ours (N=4)
& $99.74$/$99.84$ & \ $99.66$/$99.76$ & $84.84$ & $68.59$ & $77.69$ & $53.42$  \\
+Ours (N=8)
& $99.71$/$99.61$ & \ $99.61$/$99.62$ & $86.53$ & $70.38$ & $78.77$ & $55.15$  \\
\midrule
ResNet34
& $99.77$ & $99.67$ & $80.29$ &  $69.34$     & $37.35$     & $27.12$  \\
+${\cal F}$ (N=4)
& {$92.16$}/{$\textbf{62.96}$} & \ {$89.90$}/{$\textbf{60.95}$} & {$90.59$} & {$85.85$} & {$75.71$} & {$72.58$}  \\
+Ours (N=4)
& $99.74$/$99.81$ & \ $99.65$/$99.74$ & $81.72$ & $71.74$  & $54.99$ & $35.94$  \\
+Ours (N=8)
& $99.81$/$99.81$ & \ $99.74$/$99.75$ & $81.72$  & $71.74$ & $50.86$ & $36.50$   \\
\hdashline
ResNet34+${\cal K}$ 
& $99.74$ & $99.65$ &  $83.01$  &  $69.35$  & $16.73$  & $10.17$  \\
+${\cal F}$ (N=4)
& {$82.74$}/{$71.51$} & \ {$77.27$}/{$69.92$} & $83.95$ & $72.60$ & $83.43$ & $75.83$  \\
+Ours (N=4)
& $99.84$/$99.77$ & \ $99.79$/$99.69$ & $84.25$ & $71.14$ & $39.09$ & $22.93$  \\
+Ours (N=8)
& $99.68$/{$99.77$} & \ $99.56$/$99.69$ & {$84.35$} & {$72.40$}  & {$59.55$} & {$24.20$}  \\
\bottomrule
\end{tabular}
\end{adjustbox}
{\\ \footnotesize 1: test separately for using Scirep or RabbinWBC as the out-of-domain dataset.}
\end{table}

\subsubsection{Settings}
To evaluate the effectiveness of our method, we conduct experiments on two medical image classification tasks: dermatology and white blood cell (WBC) classification. For skin dataset, we train a dermatology classifier ($K = 2$) using concept annotations from the Fitzpatrick 17k dataset~\cite{groh2021evaluating} and the SkinCon dataset~\cite{daneshjou2022skincon}. We use the same concepts ($L = 22$) with~\cite{DBLP:conf/miccai/PangKTW24} to train the concept predictor. The Fitzpatrick 17k dataset ($|{\cal D}| = 3,479$) is used as the in-domain dataset and the Diverse Dermatology Images (DDI) dataset~\cite{daneshjou2022disparities} ($|{\cal D}| = 656$) serves as the out-of-domain dataset. The DDI dataset focuses on diverse skin tones and uncommon diseases, which introduces variability in skin tones, lighting conditions, and disease presentation. For the WBC dataset, we use the PBC dataset~\cite{acevedo2020dataset} combined with concept annotations ($L = 11$) from the WBCAtt dataset~\cite{tsutsui2023benchmarking}. We train the models ($K = 5$) following the data split from~\cite{tsutsui2023benchmarking}. The Scirep dataset~\cite{li2023deep} ($|{\cal D}| = 2,019$) and the RaabinWBC dataset~\cite{kouzehkanan2022large} ($|{\cal D}| = 4,339$) are used as out-of-domain datasets. These WBC datasets include images captured under different staining conditions and imaging devices, resulting in different cell appearances and textures. Example Images are shown in Fig.~\ref{datset_introduction_fig1}.

We separately use VGG16~\cite{simonyan2014very} and ResNet34~\cite{he2016deep} pre-trained on ImageNet as $g$ and use a linear classifier as $f$ to build CBMs. For both tasks, we use an AdamW optimizer~\cite{kingma2014adam} with weight decay set to 0.01. The batch size is set to 64. The learning rate is set to \texttt{1e-4}. We use cross-entropy loss with label smoothing ($\phi$=0.05) and $\varphi$ in Eq.~\ref{loss_function} is set to 1. We follow~\cite{DBLP:conf/miccai/PangKTW24} to train 30 epochs per run and also implement CBMs integrated with clinical knowledge (denoted as +${\cal K}$). Except for the additional loss functions used in~\cite{DBLP:conf/miccai/PangKTW24}, other parameters are kept the same for both vanilla CBMs and the clinical knowledge integrated CBMs. For fine-tuning experiments (denoted as + ${\cal F}$), we use the same parameters and settings with in-domain settings. The same images ($N$ for each class) are used for fine-tuning and our confusion concept identification strategy.

\subsubsection{Results}
The main results for skin images are presented in Table~\ref{tab:classification_acc_skin} and those for WBC images are in Table~\ref{tab:classification_acc_wbc}. Without any fine-tuning or fine-grained concept labels, our method can enhance the classification performance of the CBM model on out-of-domain datasets. For the vanilla CBM with VGG16 as the backbone, with only 8 images from target domain (N=4 per class), our method achieves an accuracy improvement of 4.56 and an F1 score improvement of 1.74 on the DDI dataset (bold in Table~\ref{tab:classification_acc_skin}), even surpassing out-of-domain performance when using twice the number of images for fine-tuning. More importantly, under nearly all experimental settings, our method can maintain or even improve the model's in-domain performance. In contrast, fine-tuning may lead to a drastic decline in in-domain performance. For example, the accuracy drops from 99.77 to 62.96 on WBC classification, and the F1 score drops from 99.67 to 60.95 (bold in Table~\ref{tab:classification_acc_wbc}).

\subsubsection{Are Confusion Concepts Manipulated Correctly?}
\begin{figure}[tb]
    \centering
    \includegraphics[width=\linewidth]{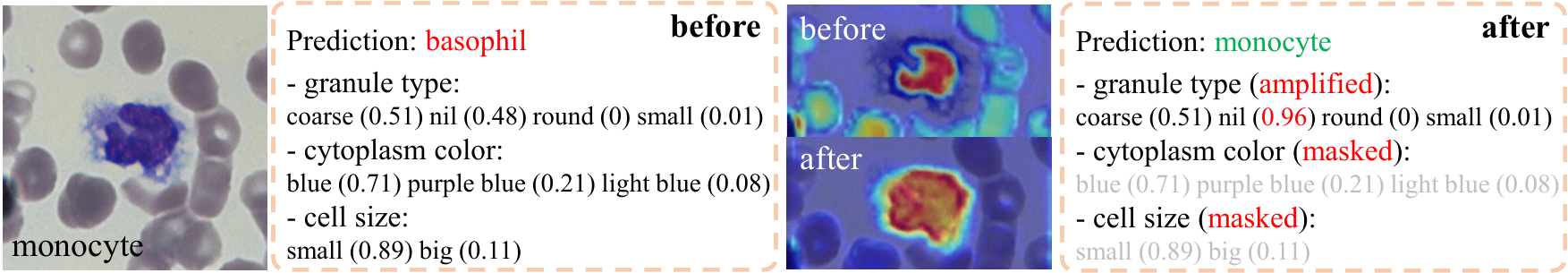}
    \caption{Visualization of the inference procedure and focus areas of the model before and after applying our test-time improvement strategy, both the classification accuracy and model faithfulness are improved.}
    \label{inference_visualization}
\end{figure}

We visualize the inference process and utilize Grad-CAM~\cite{selvaraju2020grad} to highlight the important features that the model relies on for predictions to see whether the concept manipulation is faithful. As presented in Fig.~\ref{inference_visualization}, the model becomes more focused on the target object after our confusion concept manipulation, indicating better faithfulness during inference, in addition to the improvement of classification accuracy.
\section{Conclusion}
In this study, we propose a novel training-free confusion concept identification strategy to enhance the adaptability of concept bottleneck models (CBMs) for out-of-domain data. Our approach requires only minimal image-level annotations, yet effectively improves test-time performance while maintaining in-domain accuracy and model explainability. Both quantitative evaluations and visual analyses demonstrate the efficacy of our method, revealing the potential of explainability-guided test-time adaptation methods. Future research may focus on further reducing the labeling requirement and developing label-free methods for adopting explainable medical classification models.

\begin{credits}
\subsubsection{\ackname} This work was supported by the Natural Science Foundation of China (623B2001, 82371112), the Science Foundation of Peking University Cancer Hospital (JC202505), Natural Science Foundation of Beijing Municipality (Z210008) and the Clinical Medicine Plus X - Young Scholars Project of Peking University, the Fundamental Research Funds for the Central Universities (PKU2025PKULCXQ008).

\subsubsection{\discintname}
The authors have no competing interests to declare that are relevant to the content of this article.
\end{credits}

%
%
%
\bibliographystyle{splncs04}
\bibliography{reference}

\begin{thebibliography}{10}
\providecommand{\url}[1]{\texttt{#1}}
\providecommand{\urlprefix}{URL }
\providecommand{\doi}[1]{https://doi.org/#1}

\bibitem{acevedo2020dataset}
Acevedo, A., Merino, A., Alf{\'e}rez, S., Molina, {\'A}., Bold{\'u}, L., Rodellar, J.: A dataset of microscopic peripheral blood cell images for development of automatic recognition systems. Data in brief  \textbf{30},  105474 (2020)

\bibitem{acosta2022multimodal}
Acosta, J.N., Falcone, G.J., Rajpurkar, P., Topol, E.J.: Multimodal biomedical ai. Nature Medicine  \textbf{28}(9),  1773--1784 (2022)

\bibitem{balendran2025scoping}
Balendran, A., Beji, C., Bouvier, F., Khalifa, O., Evgeniou, T., Ravaud, P., Porcher, R.: A scoping review of robustness concepts for machine learning in healthcare. npj Digital Medicine  \textbf{8}(1), ~38 (2025)

\bibitem{chen2023algorithmic}
Chen, R.J., Wang, J.J., Williamson, D.F., Chen, T.Y., Lipkova, J., Lu, M.Y., Sahai, S., Mahmood, F.: Algorithmic fairness in artificial intelligence for medicine and healthcare. Nature biomedical engineering  \textbf{7}(6),  719--742 (2023)

\bibitem{choi2025adaptive}
Choi, J., Raghuram, J., Li, Y., Jha, S.: Adaptive concept bottleneck for foundation models under distribution shifts. In: The Thirteenth International Conference on Learning Representations (2025)

\bibitem{DBLP:conf/miccai/ChowdhuryPLTXHVL24}
Chowdhury, T.F., Phan, V.M.H., Liao, K., To, M., Xie, Y., van~den Hengel, A., Verjans, J.W., Liao, Z.: Adacbm: An adaptive concept bottleneck model for explainable and accurate diagnosis. In: {MICCAI} {(10)}. Lecture Notes in Computer Science, vol. 15010, pp. 35--45. Springer (2024)

\bibitem{daneshjou2022disparities}
Daneshjou, R., Vodrahalli, K., Novoa, R.A., Jenkins, M., Liang, W., Rotemberg, V., Ko, J., Swetter, S.M., Bailey, E.E., Gevaert, O., et~al.: Disparities in dermatology ai performance on a diverse, curated clinical image set. Science advances  \textbf{8}(31),  eabq6147 (2022)

\bibitem{daneshjou2022skincon}
Daneshjou, R., Y{\"{u}}ksekg{\"{o}}n{\"{u}}l, M., Cai, Z.R., Novoa, R.A., Zou, J.Y.: Skincon: {A} skin disease dataset densely annotated by domain experts for fine-grained debugging and analysis. In: NeurIPS (2022)

\bibitem{fajtl2024trustworthy}
Fajtl, J., Welikala, R.A., Barman, S., Chambers, R., Bolter, L., Anderson, J., Olvera-Barrios, A., Shakespeare, R., Egan, C., Owen, C.G., et~al.: Trustworthy evaluation of clinical ai for analysis of medical images in diverse populations. NEJM AI  \textbf{1}(9),  AIoa2400353 (2024)

\bibitem{groh2021evaluating}
Groh, M., Harris, C., Soenksen, L., Lau, F., Han, R., Kim, A., Koochek, A., Badri, O.: Evaluating deep neural networks trained on clinical images in dermatology with the fitzpatrick 17k dataset. In: {CVPR} Workshops. pp. 1820--1828. Computer Vision Foundation / {IEEE} (2021)

\bibitem{gunning2019darpa}
Gunning, D., Aha, D.: Darpa’s explainable artificial intelligence (xai) program. AI magazine  \textbf{40}(2),  44--58 (2019)

\bibitem{he2025v2c}
He, H., Zhu, L., Zhang, X., Zeng, S., Chen, Q., Lu, Y.: V2c-cbm: Building concept bottlenecks with vision-to-concept tokenizer. In: Proceedings of the AAAI Conference on Artificial Intelligence. vol. 39(3), pp. 3401--3409 (2025)

\bibitem{he2016deep}
He, K., Zhang, X., Ren, S., Sun, J.: Deep residual learning for image recognition. In: {CVPR}. pp. 770--778. {IEEE} Computer Society (2016)

\bibitem{huang2024concept}
Huang, Q., Song, J., Hu, J., Zhang, H., Wang, Y., Song, M.: On the concept trustworthiness in concept bottleneck models. In: {AAAI}. pp. 21161--21168. {AAAI} Press (2024)

\bibitem{DBLP:conf/miccai/KimKCK23}
Kim, I., Kim, J., Choi, J., Kim, H.J.: Concept bottleneck with visual concept filtering for explainable medical image classification. In: ISIC/Care-AI/MedAGI/DeCaF@MICCAI. Lecture Notes in Computer Science, vol. 14393, pp. 225--233. Springer (2023)

\bibitem{kingma2014adam}
Kingma, D.P.: Adam: A method for stochastic optimization. arXiv preprint arXiv:1412.6980  (2014)

\bibitem{DBLP:conf/icml/KohNTMPKL20}
Koh, P.W., Nguyen, T., Tang, Y.S., Mussmann, S., Pierson, E., Kim, B., Liang, P.: Concept bottleneck models. In: {ICML}. Proceedings of Machine Learning Research, vol.~119, pp. 5338--5348. {PMLR} (2020)

\bibitem{kouzehkanan2022large}
Kouzehkanan, Z.M., Saghari, S., Tavakoli, S., Rostami, P., Abaszadeh, M., Mirzadeh, F., Satlsar, E.S., Gheidishahran, M., Gorgi, F., Mohammadi, S., et~al.: A large dataset of white blood cells containing cell locations and types, along with segmented nuclei and cytoplasm. Scientific reports  \textbf{12}(1), ~1123 (2022)

\bibitem{li2023deep}
Li, M., Lin, C., Ge, P., Li, L., Song, S., Zhang, H., Lu, L., Liu, X., Zheng, F., Zhang, S., et~al.: A deep learning model for detection of leukocytes under various interference factors. Scientific Reports  \textbf{13}(1), ~2160 (2023)

\bibitem{DBLP:journals/ijcv/LiangHT25}
Liang, J., He, R., Tan, T.: A comprehensive survey on test-time adaptation under distribution shifts. Int. J. Comput. Vis.  \textbf{133}(1),  31--64 (2025)

\bibitem{DBLP:conf/nips/MaiC00CPB0S0C24}
Mai, Z., Chowdhury, A., Zhang, P., Tu, C., Chen, H., Pahuja, V., Berger{-}Wolf, T.Y., Gao, S., Stewart, C.V., Su, Y., Chao, W.: Fine-tuning is fine, if calibrated. In: NeurIPS (2024)

\bibitem{mccloskey1989catastrophic}
McCloskey, M., Cohen, N.J.: Catastrophic interference in connectionist networks: The sequential learning problem. In: Psychology of learning and motivation, vol.~24, pp. 109--165. Elsevier (1989)

\bibitem{moor2023foundation}
Moor, M., Banerjee, O., Abad, Z.S.H., Krumholz, H.M., Leskovec, J., Topol, E.J., Rajpurkar, P.: Foundation models for generalist medical artificial intelligence. Nature  \textbf{616}(7956),  259--265 (2023)

\bibitem{DBLP:conf/iclr/OikarinenDNW23}
Oikarinen, T.P., Das, S., Nguyen, L.M., Weng, T.: Label-free concept bottleneck models. In: {ICLR}. OpenReview.net (2023)

\bibitem{DBLP:conf/miccai/PangKTW24}
Pang, W., Ke, X., Tsutsui, S., Wen, B.: Integrating clinical knowledge into concept bottleneck models. In: {MICCAI} {(4)}. Lecture Notes in Computer Science, vol. 15004, pp. 243--253. Springer (2024)

\bibitem{accuracy_not_fairness_NEJM}
Sabuncu, M.R., Wang, A.Q., Nguyen, M.: Ethical use of artificial intelligence in medical diagnostics demands a focus on accuracy, not fairness. NEJM AI  \textbf{2}(1),  AIp2400672 (2025)

\bibitem{selvaraju2020grad}
Selvaraju, R.R., Cogswell, M., Das, A., Vedantam, R., Parikh, D., Batra, D.: Grad-cam: Visual explanations from deep networks via gradient-based localization. Int. J. Comput. Vis.  \textbf{128}(2),  336--359 (2020)

\bibitem{DBLP:conf/icml/ShinJAL23}
Shin, S., Jo, Y., Ahn, S., Lee, N.: A closer look at the intervention procedure of concept bottleneck models. In: {ICML}. Proceedings of Machine Learning Research, vol.~202, pp. 31504--31520. {PMLR} (2023)

\bibitem{simonyan2014very}
Simonyan, K., Zisserman, A.: Very deep convolutional networks for large-scale image recognition. arXiv preprint arXiv:1409.1556  (2014)

\bibitem{DBLP:conf/icml/SteinmannSFK24}
Steinmann, D., Stammer, W., Friedrich, F., Kersting, K.: Learning to intervene on concept bottlenecks. In: {ICML}. OpenReview.net (2024)

\bibitem{tsutsui2023benchmarking}
Tsutsui, S., Su, Z., Wen, B.: Benchmarking white blood cell classification under domain shift. In: {ICASSP}. pp.~1--5. {IEEE} (2023)

\bibitem{Varghese2024aiinsurgery}
Varghese, C., Harrison, E.M., O'Grady, G., Topol, E.J.: Artificial intelligence in surgery. Nature medicine  \textbf{30}(1),  1257--1268 (2024)

\bibitem{DBLP:conf/nips/0006GWWYCGY24}
Yang, Y., Gandhi, M., Wang, Y., Wu, Y., Yao, M.S., Callison{-}Burch, C., Gee, J.C., Yatskar, M.: A textbook remedy for domain shifts: Knowledge priors for medical image analysis. In: NeurIPS (2024)

\bibitem{DBLP:conf/cvpr/YangPZJCY23}
Yang, Y., Panagopoulou, A., Zhou, S., Jin, D., Callison{-}Burch, C., Yatskar, M.: Language in a bottle: Language model guided concept bottlenecks for interpretable image classification. In: {CVPR}. pp. 19187--19197. {IEEE} (2023)

\bibitem{kang2025aiindrug}
Zhang, K., Yang, X., Wang, Y., Yu, Y., Huang, N., Li, G., Li, X., Wu, J.C., Yang, S.: Artificial intelligence in drug development. Nature medicine  \textbf{31}(1),  45--59 (2025)

\end{thebibliography}
\nolinenumbers
\end{document}